\documentclass[lettersize,journal]{IEEEtran}
\usepackage{amsmath,amsfonts}
\usepackage{algorithmic}
\usepackage{algorithm}
\usepackage{array}
\usepackage[caption=false,font=normalsize,labelfont=sf,textfont=sf]{subfig}
\usepackage{textcomp}
\usepackage{stfloats}
\usepackage{url}
\usepackage{verbatim}
\usepackage{graphicx}
\usepackage{cite}
\usepackage{xcolor}
\usepackage{cprotect}
\newcommand{\OurMethod}{{AIR-DA}}
\newcommand{\etal}{\textit{et al}. }
\newcommand{\ie}{\textit{i}.\textit{e}., }

\begin{document}

\title{AIR-DA: Adversarial Image Reconstruction for Unsupervised Domain Adaptive Object Detection}

\author{Kunyang Sun,  Wei Lin, Haoqin Shi, Zhengming Zhang, Yongming Huang, and Horst Bischof}



\maketitle

\begin{abstract}
Unsupervised domain adaptive object detection is a challenging vision task where object detectors are adapted from a label-rich source domain to an unlabeled target domain. Recent advances prove the efficacy of the adversarial based domain alignment where the adversarial training between the feature extractor and domain discriminator results in domain-invariance in the feature space.
However, due to the domain shift, domain discrimination, especially on low-level features, is an easy task. This results in an imbalance of the adversarial training between the domain discriminator and the feature extractor.
In this work, we achieve a better domain alignment by introducing an auxiliary regularization task to improve the training balance.
Specifically, we propose Adversarial Image Reconstruction (AIR) as the regularizer to facilitate the adversarial training of the feature extractor.
We further design a multi-level feature alignment module to enhance the adaptation performance. 
Our evaluations across several datasets of challenging domain shifts demonstrate that the proposed method outperforms all previous methods, of both one- and two-stage, in most settings.
\end{abstract}


\section{Introduction}
\renewcommand{\thefootnote}{}
\footnotetext{
This work was done during the research stay of K. Sun at the Institute of Computer Graphics and
Vision, Graz University of Technology.

K. Sun, H. Shi, Z. Zhang and Y. Huang are with the School of Information Science and Engineering, the National Mobile Communications Research Laboratory, Southeast University, Nanjing 210096, China. H. Shi, Z. Zhang and Y. Huang are also with the Pervasive Communications Center, Purple Mountain Laboratories, Nanjing 211111, China (e-mail: \{Sunky, shihaoqing619, zmzhang, huangym\}@seu.edu.cn).

W. Lin and H. Bischof are with the Institute for Computer Graphics and Vision, Graz University of Technology, Graz 8010, Austria
(e-mail: \{wei.lin, bischof\}@icg.tugraz.at).}
Object detection, with recognition and localization of object instances in visual data, is one of the fundamental computer vision tasks.
Due to the rapid development of deep convolutional neural networks (DCNN)~\cite{res101,vgg16}, 
object detection has achieved tremendous success in recent years~\cite{FCOS,SSD,lin2017focal,dai2017deformable,cai2018cascade}, with the emergence of top-performing models such as Faster R-CNN~\cite{frcnn} and YOLO~\cite{yolo,redmon2018yolov3,bochkovskiy2020yolov4}.
However, in real world scenarios, the detector trained in one environment has degraded performance when applied on data captured in the other, due to the domain shift such as different camera viewpoints, styles of object appearances and weather conditions. 

Therefore, unsupervised domain adaptation (UDA) approaches are proposed to bridge the domain gap, adapting the deep models from an annotated source domain to an unlabeled target domain~\cite{xie2019,roy2019,he2020,khodabandeh2019,zhuang2020,li2020free,guan2021uncertainty,zhang2018single}.
A typical crowd of UDA methods are based on adversarial training via the Gradient Reversal Layer (GRL), between a feature extractor and a domain discriminator with regard to a domain discrimination task. These adversarial-based approaches are extensively evaluated on image classification~\cite{uda}, semantic segmentation~\cite{da_semantic}, and also, object detection~\cite{da_frcnn,EPM,ICR-CCR}.

\begin{figure}[!t]

\includegraphics[scale=0.265]{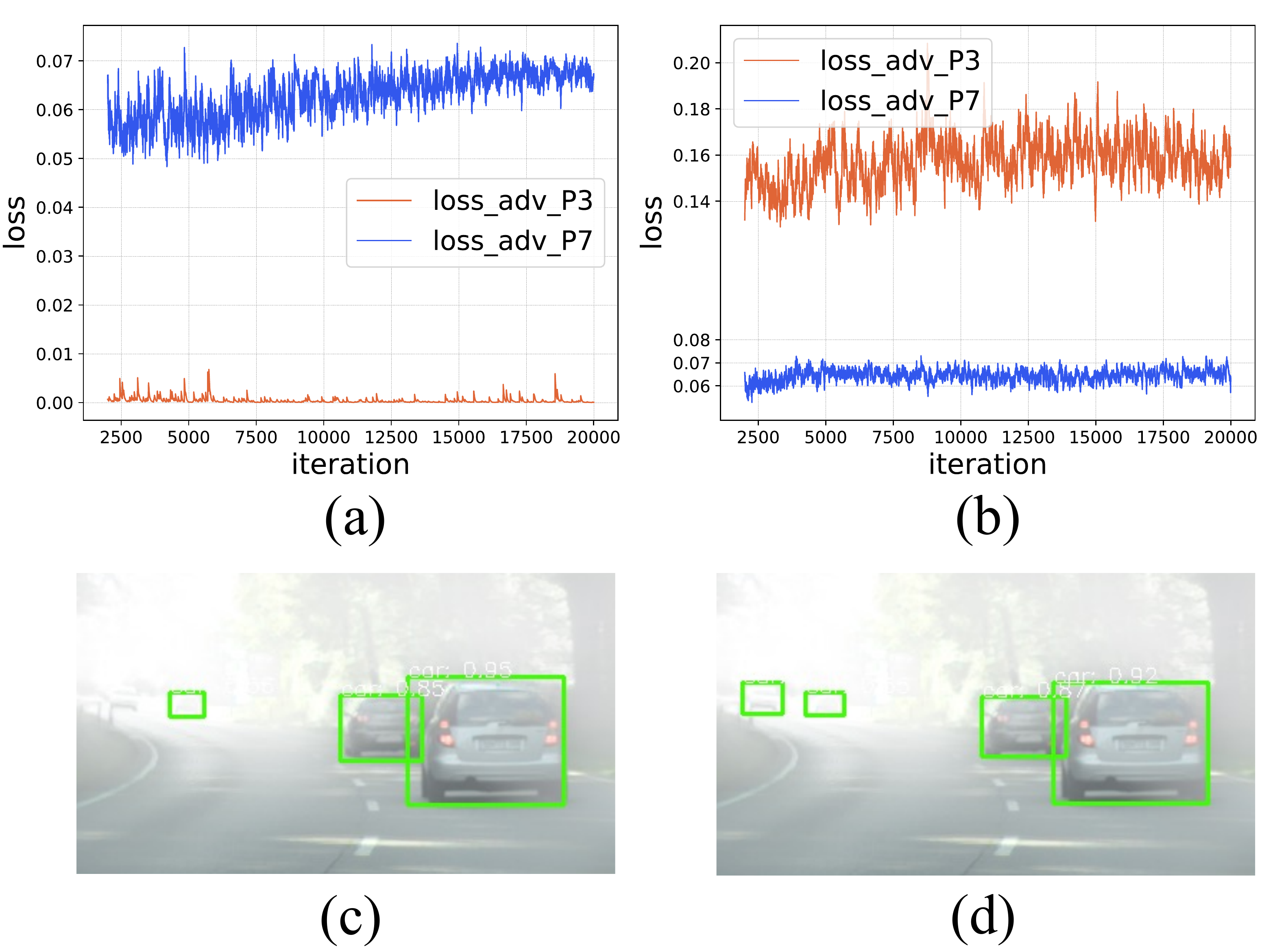}
 \cprotect\caption{We illustrate the adversarial training loss with~\OurMethod~ (subfigure (b)) and without (subfigure (a)) on a low-level feature map $\mathbf P_3$ (denoted as \verb|loss_adv_P3|) in FPN~\cite{FPN} layers. In (a) and (b), we also plot the adversarial training loss without~\OurMethod~on a high-level feature map $\mathbf P_7$ (denoted as \verb|loss_adv_P7|)) as a reference. We visualize the detection results with~\OurMethod~(subfigure (d)) and without (subfigure (c)) correspondingly. Clearly, the trend of loss on \verb|loss_adv_P3| demonstrates that the proposed AIR module improves the balance of adversarial training on the low-level feature, which leads to enhanced detection performance.}
\label{fig:intro}
\end{figure}

Domain Adaptive Faster R-CNN~\cite{da_frcnn}, as a seminal work of adversarial-based domain alignment for object detection, is based on the two-stage detector R-CNN~\cite{frcnn}.
This work performs domain alignment on both image-level and instance-level, by aligning features of the backbone and the region proposals respectively.
However, in spite of the simple architecture, it has two main drawbacks:
(1) The detector trained on source images only is unaware of the target distribution. Therefore, the predictions of region proposals on target images are inaccurate and contain excessive irrelevant background contents. Consequently, both image- and instance-level domain alignment dominated by background contents lead to further error propagation. (2) Due to the existent domain shift, the domain discrimination loss quickly approaches zero and there is minimum gradient flow. This results in an imbalance of the adversarial training between the domain discriminator and the feature extractor, leading to inferior adaptation performance.

To address the first drawback, Zhu \etal~\cite{SC-DA} use unsupervised clustering to adjust the alignment region, while Saito \etal~\cite{SW-DA} employ the focal loss to force the network to focus on foreground contents that are more likely to be domain-invariant.
This idea has been explored by EPM~\cite{EPM}, which proposes a pixel-level alignment method based on the one-stage detector FCOS~\cite{FCOS}. 
The center-ness branch of FCOS generates attention maps to focus on positive samples.

Despite the efforts on tackling the first drawback, the solutions to the second drawback of training imbalance are still underdeveloped, only with attempts of multi-level domain alignments~\cite{MAF,EPM}. Specifically, He \etal~\cite{MAF} attempt to exploit a multi-level domain adaptation model for aligning both low- and high-level features. Similarly, Hsu \etal~\cite{EPM} also employ an attention map in multi-level alignments for better performance. However, due to the large domain shift in low-level features, there is an imbalance between training of the feature extractor and the domain discriminator. Therefore, merely performing adversarial domain discrimination fails to deliver satisfying alignment performance. 



Rethinking domain adaptation on classification problems, where the classification task is only performed on the high-level features, merely aligning high-level features which contain rich semantic information can already achieve satisfactory performance. The imbalance in adversarial training has not drawn much attention. However, in the case of object detection, where detection heads are applied on features of multiple levels, multi-level domain alignment is essential for tackling the unsupervised domain adaptive object detection (UDA Object Detection).

Furthermore, compared to high-level features, low-level features with domain-specific fine-grained details demonstrate larger domain shift and are more easily distinguished by a domain discriminator. This causes the imbalance in the adversarial training. As shown in Figure~\ref{fig:intro} (a), only using a domain discriminator, the adversarial loss of the high-level feature maps (P7) demonstrates some balance. However, the loss on the low-level feature maps (P3) rapidly approaches zero. This results in minimum gradient flow for adversarial training of the feature extractor, leading to inferior domain alignment.

In this context, we propose the novel method of Adversarial Image Reconstruction for Domain Adaptation (AIR-DA) as a regularization to achieve better adversarial training balance. 

We attend to the interesting question: \textit{With the presence of large domain shift, how do we make the domain discrimination task more challenging for the discriminator?} We argue that the key is to eliminate the fine-grained details in the features. To this end, inspired by the image reconstruction task that depend on fine-level details, we propose the branch of Adversarial Image Reconstruction (AIR) as a regularization. By impeding the reconstruction, it encourages the feature extractor to focus on more generalized coarse-level contents of images. The elimination of fine-grained details complicates the domain discrimination task for the discriminator and thus improves the balance in the adversarial training. Moreover, AIR is a lightweight reconstruction head added after the GRL to the feature extractor, which is removed during the inference. As shown in Figure~\ref{fig:intro} (b) and (d), the proposed AIR branch improves the balance of adversarial training and leads to better detection performance.

We further extend the AIR branch to a multi-level alignment module. Different than related work that simply applies the domain alignment module feature of each layer, we propose different AIR modules for low- and high-level features separately. Specifically, we pass low-level features directly into the image reconstruction branch. In the meanwhile, we fuse high-level features with backbone features and feed the fusion result for reconstruction. Our ablation studies validate the design of this multi-level extension.

We perform extensive evaluations on scenarios of various domain shifts, \ie weather variation ( Cityscapes~\cite{cityscape} to Foggy Cityscapes~\cite{foggy-cityscape}), cross-camera adaptation ( KITTI~\cite{kitti} to Cityscapes~\cite{cityscape}) as well as synthetic-to-real-world adaptation (Sim10k~\cite{sim10k} to Cityscapes~\cite{cityscape}). The proposed \OurMethod{}, applied on the one-stage object detector FCOS~\cite{FCOS}, outperforms UDA Object Detection state-of-the-arts of both one- and two-stage architecture.

Our main contributions are as follows:
\begin{itemize}
    
    \item  We propose a lightweight and efficient framework to address the imbalance of adversarial training between feature extractors and domain discriminators. To the best of our knowledge, \OurMethod~is the first to utilize the adversarial image reconstruction (AIR) as a self-supervised auxiliary task for tackling the imbalance in adversarial training of domain discrimination.
    
    
     \item  We extend the AIR branch to a multi-level alignment module by proposing different AIR modules for low- and high-level features separately.

    \item  Our extensive evaluation across various domain shift scenarios validate the efficacy of the proposed framework. Without bells and whistles, Our~\OurMethod,  applied on the one-stage detector FCOS, outperforms both one- and two-stage UDA Object Detection state-of-the-arts.

\end{itemize}

\section{Related Work}

\begin{figure*}[!t]
\centerline{\includegraphics[scale=0.66]{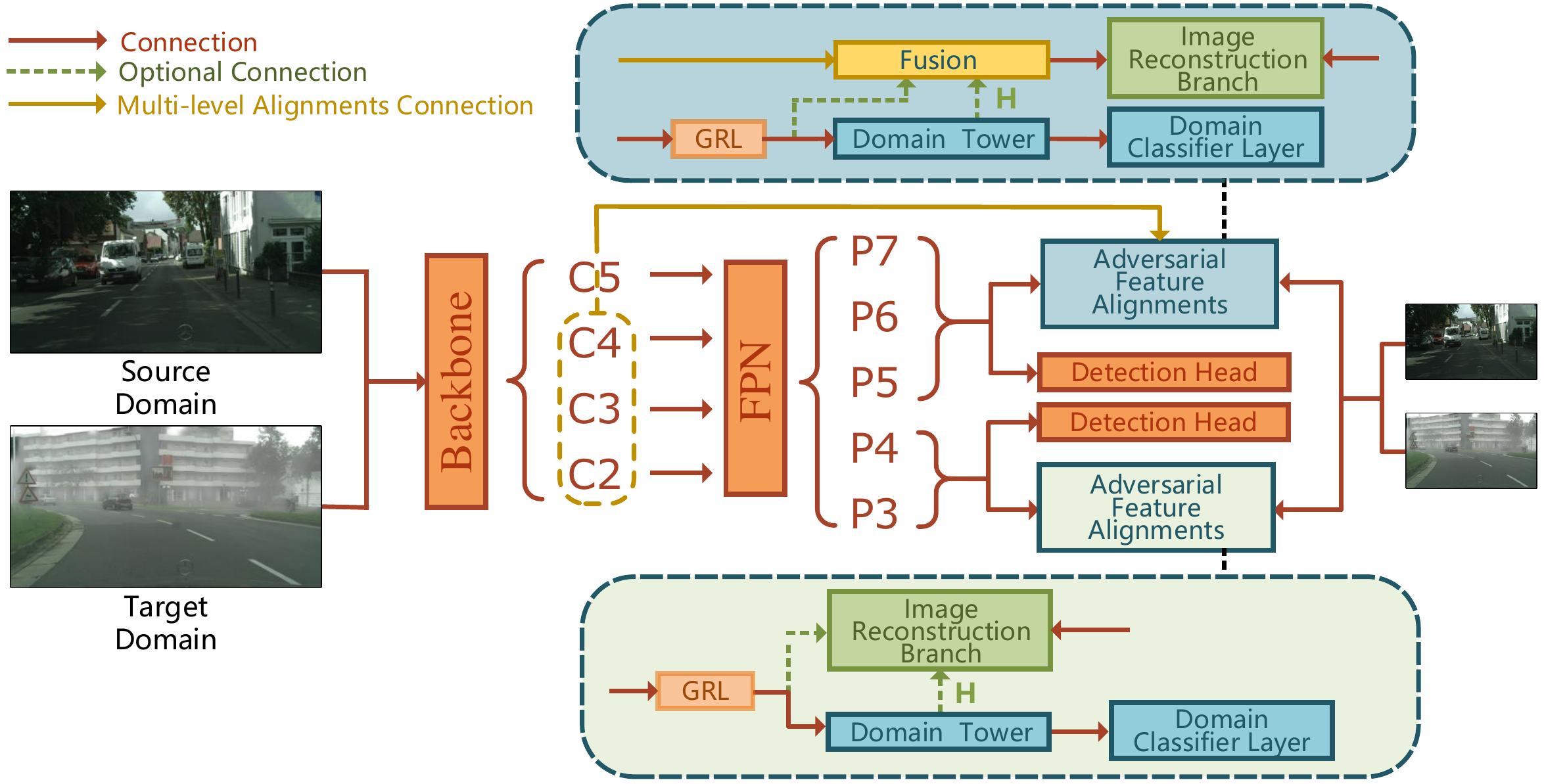}}
\caption{ \footnotesize \textbf{\OurMethod{} pipeline} Built on the powerful one-stage object detector, our adversarial feature alignment module includes two parts: domain discriminator and adversarial image reconstruction (AIR) module. The domain discriminator consists of (1) a Gradient Reversal Layer (GRL)~\cite{uda} to reverse backward gradients on the feature extractor (2) a domain tower for studying domain features and (3) a classification layer to identify the input domain. The AIR module exploits either the output of the domain tower ($\bf H$) or the FPN features after GRL to reconstruct the input image. Note that the input of high-level AIR module fuses backbone features that are passed through a GRL.}
\label{fig:main}
\end{figure*}

\subsection{Object Detection}
\textbf{Anchor-based detectors} Similar to the traditional sliding-window, the pre-defined anchors can be categorized into object proposals or background patches. In those detectors, although one-stage detectors like SSD~\cite{SSD} win at speed, top-performing systems are also of a two-stage pipeline~\cite{frcnn}. They apply a region proposal network (RPN) to predict regions of interest (RoIs) and attach a set of branches on the RoI features for final predictions. 

\textbf{Anchor-free detectors} Benefiting from the fully convolutional network (FCN), many detectors generate dense predictions for keypoint like center~\cite{centernet} or corners~\cite{CornerNet} of boxes and use some techniques to group them. While they could get rid of the region proposal module, these works are suffering from heavy post-processing. Tian~\textit{et al.} propose FCOS~\cite{FCOS} which follows a per-pixel prediction paradigm. It utilizes FPN~\cite{FPN} to learn objects in different scales and introduces center-ness to suppress low-quality prediction samples. In this work, we adopt FCOS as our detector for the UDA Object Detection task, due to its competence and efficiency.
\subsection{Domain Adaptive Object Detectors}
\textbf{Two-stage DA detectors}  DADO approaches are dominated by two-stage methodologies.
Thanks to the Gradient Reversal Layer (GRL)~\cite{uda} and the powerful detector Faster R-CNN~\cite{frcnn}, Chen \textit{et al.} propose DA Faster R-CNN~\cite{da_frcnn}, which aligns both image-level and instance-level features by adversarial learning separately. For instance-level alignment, SC-DA~\cite{SC-DA} inherits the ideas from k-means clustering and improves DA Faster R-CNN by aggregating region proposals for higher quality instance regions. However, clustering also brings new hyper-parameters which need to be carefully tuned. 

Saito \textit{et al.} claim that strong matching of low-level features is more meaningful and propose SW-DA~\cite{SW-DA} by combining strong local feature alignment with weak global feature alignment. However, as low-level features have a large domain gap, this method suffers from the training imbalance between domain discriminator and feature extractor. Even though SW-DA are further enhanced by ICR-CCR~\cite{ICR-CCR} that leverages weak localization ability of convolutional neural network (CNN) for multi-label classification to force network aligning object-related local regions, there is no solution for the above issue. 

\textbf{Single-stage DA detectors} Based on the effective detector SSD, Kim \textit{et al.} present a weak self-training method that can reduce negative effects of inaccurate pseudo-labels~\cite{kimssd}. Benefiting from the effectiveness from FCOS~\cite{FCOS}, EPM~\cite{EPM} is proposed as a pixel-level alignment method. Specifically, they first apply multi-level adversarial alignments on corresponding FPN feature maps. Then, they use center-ness prediction to discover center of object regions and focus on aligning them. However, it is a two-step training method as it requires training global-aware (GA) alignments before center-aware (CA) ones. Moreover, as EPM is sensitive to the pretrained model for the CA alignment training session, careful finetuning of GA training is required. Above methods do not attend to the imbalance of adversarial training in feature alignments, especially in local features. To date, almost all such approaches fall behind two-stage methods in prediction precision, leaving significant room for improvement.

To break these limitations, we propose a new one-stage domain adaptive detector, termed \OurMethod. We upgrade the adversarial structure as well as balance the adversarial training by introducing a novel AIR module. The domain discrimination loss (Figure~\ref{fig:intro}) indicates that, compared to adaptation with domain discriminators only, the proposed algorithm has better training balance. Without any doubt, as shown in our experiments (Section \ref{Experiments}), our \OurMethod~ also improves the performance on local feature alignments.

\subsection{Image reconstruction for UDA}
Image reconstruction is a classic self-supervised task. The input image is processed by the feature extractor and then recovered from the reconstruction network. In domain adaptation for classification task, many previous works also employ reconstruction strategies. Bousmalis~\textit{et al.}~\cite{DSN} apply a shared encoder and two private encoders to separate and save domain-invariant features. The image reconstruction loss is minimizedure with the goal of ensuring completeness and validity of both shared and private features. Ghifary~\textit{et al.}~\cite{DRCNs} consider reconstructing target domain images to push the backbone network to study target domain features. However, none of them focuses on the training imbalance problem or trains the image reconstruction task in an adversarial fashion. It is worth noting that our \OurMethod~ employs the proposed adversarial image reconstruction branch as a regularization for  improving balance of the adversarial training of domain discrimination. 
Our evaluation across datasets of various domain shifts demonstrate the advanced detection performance, benefiting from an improved training balance.

\section{Our Approach}
\subsection{Framework overview}
The overview of our \OurMethod{} framework is illustrated in Figure~\ref{fig:main}. It is composed of a detector network and an adversarial feature alignments module. Notably, both of them share the same feature extractor and are trained in an end-to-end fashion. 
As source images $\mathbf{I}_{s}$ have ground truth bounding box labels $\mathbf{B}_{s}$ which are lacked in target images $\mathbf{I}_{t}$, the detector network are trained by $\mathbf{I}_{s}$. The adversarial feature alignments module has two parts, a domain discrimination net to classify the input domain and a reconstruction branch to restore input images. Inspired by FCOS~\cite{FCOS} which works in a multi-level prediction manner by FPN~\cite{FPN},~\OurMethod{} also applies multi-level adversarial feature alignments for adapting different FPN layers.

\subsection{Domain discriminator for feature alignments}
Following the adversarial alignment techniques in DA Faster R-CNN series~\cite{da_frcnn}, we apply a domain discrimination network to classify the domain at each location on input feature maps. In particular, the feature map $\mathbf P$, as the output from the feature extractor, has a shape of $N\times K\times \frac H s \times \frac W s$, where $N$ is the batch size, $K$ is the channel dimension, $H\times W$ is the input image size, and $s$ is the feature map output stride. First, we apply a GRL~\cite{uda} to reverse the backward gradient w.r.t the domain discrimination loss in backbone and FPN. Then we  contruct a light CNN with four $3\times 3$ convolutional layers as the domain tower and a single classification layer to predict a domain discrimination map $\mathbf A$ with the shape of $N\times \frac H s \times \frac W s$, indicating the domain of each corresponding location on the feature map $\mathbf P$. Here we denote $\mathbf H\in \mathbb R_{>0}^{N\times K\times \frac H s \times \frac W s}$ as the output of the domain tower. Following other adversarial detectors, we set the domain label $d$ to 1 for the source domain and 0 for the target domain. Then, the adversarial domain loss is defined as follows:
\begin{equation}\label{eq:1}
    \mathcal{L}_{ad} = -\sum_{u,v}\Big[ d\cdot\mathrm{log} (\sigma(\mathbf{A}^{(u,v)})) + (1-d)\mathrm{log} (1-\sigma(\mathbf{A}^{(u,v)})) \Big],
\end{equation}
where $\sigma $ denotes the $sigmoid$ function.

\begin{figure}[t]
\includegraphics[scale=0.75]{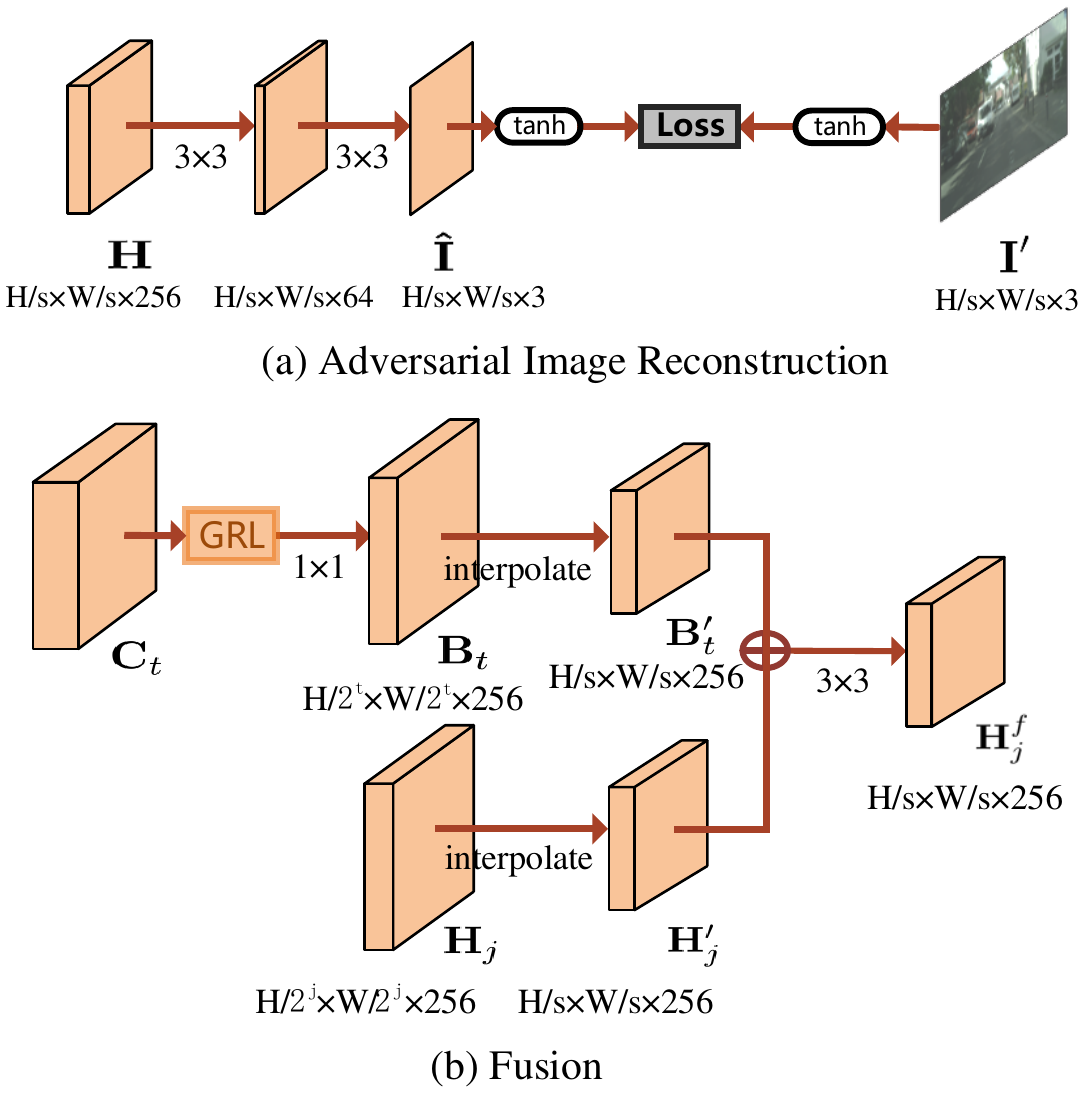}
\caption{ \footnotesize AIR module in our method. Figure (a) represents the process of adversarial image reconstruction branch $\mathcal R$. In (a), $\mathbf H$ represents the output of domain tower while $\mathbf{\hat{I}}$ are defined as recovered images of our AIR branch. Figure (b) represents the process of fusion. In (b), $\mathbf C_t$ are backbone feature maps. Besides, $\mathbf H_j$ means outputs of domain tower in high-level features and $\mathbf H^f_j$ means high-level fusion feature maps. For high-level features, they first go through Figure (b) part and then enter Figure (a) session. 
$s$ means output stride.}
\label{figure:AIR module}
\end{figure}

\subsection{Adversarial image reconstruction (AIR) module}
As mentioned before, existing methods may be limited by the imbalance of adversarial training between the domain discrimination branch and the feature extractor. To overcome this shortcoming, we propose the AIR module $\mathcal R$. More specifically, we add two $3\times 3$ convolution layers after the domain tower and process its output $\mathbf H$ by $\mathcal R$ to generate reconstruction prediction $\mathbf{\hat{I}}$. It is noteworthy that our branch $\mathcal R$ can also be attached after the GRL~\cite{uda}, paralleling with the domain tower. The green dotted line in Figure~\ref{fig:main} shows this choice. The shape of $\mathbf{\hat{I}}$ is $N\times 3 \times \frac H s \times \frac W s$ where $3$ is the dimension of the RGB channels of the image. As the reconstruction size $\frac H s \times \frac W s$ is smaller than the input images $\mathbf I$ of the whole framework, we downsample $\mathbf I$ by bilinear interpolation to the size of the reconstruction $\mathbf{\hat{I}}$ for ease of computation,
\begin{equation}\label{eq:2}
    \mathbf I' = \mbox{interpolate}_{H\times W\rightarrow \frac H s\times \frac H s}(\mathbf I).
\end{equation}
Then, both $\mathbf I'$ and $\mathbf{\hat{I}}$ are normalized to the range of $[-1,1]$ with the tanh function:
\begin{equation}\label{eq:3}
    \mathbf I'_d = \mbox{tanh}(\mathbf I'),~~ \mathbf{\hat{I}}_d = \mbox{tanh}(\mathbf{\hat{I}}).
\end{equation}
The branch $\mathcal R$ can be optimized by minimizing the adversarial reconstruction loss as follows:
\begin{equation}\label{eq:4}
    \mathcal{L}_{are} =\frac{1}{N}\sum_{i=1}^N \Vert{\mathbf I'_{d,i}-\mathbf{\hat{I}}_{d,i}}\Vert_{F}^{2},
\end{equation}
where $\Vert\cdot \Vert_{F}^{2}$ is the squared Frobenius norm. We point out that $\mathcal{L}_{are}$ needs to be computed both in source and target domain. Moreover, the AIR branch for high-level features is designed to merge backbone features and we will give details in the following subsection.

\subsection{Multi-level alignments with FPN}
Since FPN~\cite{FPN} is exploited to detect objects in different scales in FCOS~\cite{FCOS}, we believe that designing a multi-level feature alignment module can provide better alignments for DAOD. Following \cite{EPM}, for each pyramid layer, we add a domain discriminator separately. Therefore, for each FPN feature map $\mathbf P_j$ where $j \in \{3,4,5,6,7\}$, we have a corresponding $\mathbf H_j$. 
The low-level feature map $\mathbf P_3$, with the maximum resolution in FPN, preserves most fine-grained details as well as information of the small objects. 
However, high-level feature maps like $\mathbf P_6$ and $\mathbf P_7$ have smaller sizes, containing less local information. Therefore, reconstructing image from high-level features is more difficult than from low-level features of large resolution. Recognizing these difficulties, we utilize multi-scale fusion that using outputs of backbone block $\mathbf {C_t}$ where $t \in \{2,3,4\}$ to make up for the deficiency. In particular, as shown in Figure \ref{figure:AIR module} (b), for each $\mathbf C_t$, we first pass it through a GRL~\cite{uda}. Then, the output is fed into a single $1\times 1$ bottleneck layer, resulting in $\mathbf B_t$ with the shape of $N\times K\times \frac H {2^{t} }\times \frac W {2^{t} }$. We further rescale $\mathbf B_t,~t \in \{2,3,4\}$ and $\mathbf H_j,~j \in \{5,6,7\}$ by interpolation into the same size as $\mathbf H_5$
\begin{equation}\label{eq:5}
\begin{array}{l}
\mathbf B'_t = \mbox{interpolate}_{{\frac H {2^{t}}}\times {\frac W {2^{t}}}\rightarrow \frac H s\times \frac H s}(\mathbf B_t),\\
\mathbf H'_j = \mbox{interpolate}_{{\frac H {2^{j}}}\times {\frac W {2^{j}}}\rightarrow \frac H s\times \frac H s}(\mathbf H_j),\\
\end{array}
\end{equation}
where $s = 2^{5}$. Then, $\mathbf H'_j,~j \in \{5,6,7\}$ are fused with $\mathbf B'_{9-j}$,  For example, $\mathbf H'_5$ combines with $\mathbf B'_4$, $\mathbf H'_6$ combines with $\mathbf B'_3$ and so on
\begin{equation}\label{eq:6}
\begin{array}{l}
\mathbf h^f_j = \mathbf H'_j + \mathbf B'_{9-j}.
\end{array}
\end{equation}
The fusion $\mathbf h^f_j$ is passed into a single $3\times 3$ convolutional layer, resulting in $\mathbf H^f_j,~j\in \{5,6,7\}$. We further feed $\{ \mathbf H_3,\mathbf H_4,\mathbf H^f_5,\mathbf H^f_6,\mathbf H^f_7\}$ into the corresponding AIR branch. In our experiments, we also find that for a shallow backbone like VGG-16, applying AIR on $\mathbf H_3, \mathbf H_4$ already leads to great performance.

\textbf{Loss function} Since the target domain images are unlabeled, we use $\mathcal{L}_{det}$ to denote the detection training loss, which are only applied on source data. Therefore, the overall objective $\mathcal{L}_{DAE}$ for our domain adaptive detector can be written as
\begin{equation}\label{eq:7}
\begin{array}{l}
\mathcal{L}_{DAE} = \mathcal{L}_{det} +  \sum_{i}(\lambda \mathcal{L}_{ad,F_i} + \beta \mathcal{L}_{are,F_i}),
\end{array}
\end{equation}
where $\mathcal{L}_{ad,i}$ and $\mathcal{L}_{are,i}$ are defined as the adversarial domain loss and the adversarial reconstruction loss of the $i$-th FPN layer while  $\lambda$ and $\beta$ are hyper-parameters for a trade-off between the two parts of detection and domain alignment.

\begin{table*}[t]
\centering
\small
\caption{ \footnotesize Quantitative results for Cityscapes $\rightarrow$ Foggy-Cityscapes dataset. Results of each category are evaluated on mAP$^r_{0.5}$}
\resizebox{\textwidth}{!}{
\begin{tabular}{rc c c c c c c c c|c}
\hline
Methods&Backbone&person&rider& car& truck& bus &train &mbike & bicycle & mAP$^r_{0.5}$\\
\hline
\hline
Faster R-CNN (source-only) & &24.4 &30.5 &32.6 &10.8 &25.4 & 9.1 &15.2 &28.3 &22.0\\
DA-Faster~\cite{da_frcnn}& &25.0 &31.0 &40.5 &22.1 &35.3 &20.2 &20.0 &27.1 &27.6\\
SC-DA~\cite{SC-DA}& &33.5 &38.0 &48.5 &26.5 &39.0 &23.3 &28.0 &33.6 &33.8\\
MAF~\cite{MAF}& &28.2 &39.5 &43.9 &23.8 &39.9 &33.3 &29.2 &33.9 &34.0\\
SW-DA~\cite{SW-DA}& &29.9 &42.3 &43.5 &24.5 &36.2 &32.6 &30.0 &35.3 &34.3\\
D\&Match~\cite{Dmatch}& &30.8 &40.5 &44.3 &27.2 &38.4 &34.5 &28.4 &32.2 &34.6\\
MTOR~\cite{MTOR}&VGG-16 &30.6 &41.4 &44.0 &21.9 &38.6 &40.6 &28.3 &35.6 &35.1\\
EPM~\cite{EPM}& &41.9 &38.7 &56.7 &22.6 &41.5 &26.8 &24.6 &35.5 &36.0\\
CDN~\cite{CDN}& &35.8 &45.7 &50.9 &30.1 &42.5 &29.8 &30.8 &36.5 &36.6\\
ICR-CCR~\cite{ICR-CCR}& &32.9 &43.8 &49.2 &27.2 &\textbf{45.1} &36.4 &30.3 &34.6 &37.4\\
ATF~\cite{ATF}& &34.6 &47.0 &50.0 &23.7 &43.3 &38.7 &33.4 &\textbf{38.8} &38.7\\
MCAR~\cite{MCAR}& &32.0 &42.1 &43.9 &\textbf{31.3} &44.1 &43.4 &37.4 &36.6 &38.8\\
RPA~\cite{RPA}& &33.3 &45.6&50.5 &30.4 &43.6 &42.0 &29.7 &36.8 &39.0\\
Prior-DA~\cite{Prior-DA}& &36.4 &\textbf{47.3}&51.7 &22.8 &47.6 &34.1 &\textbf{36.0} &38.7 &39.3\\
FCOS (source-only) & &30.5 &23.9 &34.2 &5.8 &11.1 &5.1 &10.6 &26.1 &18.4\\
\emph{DC-only} FCOS & &38.7 &36.1 &53.1 &21.9 &35.4 &25.7 &20.6 &33.9 &33.2\\
Ours & &\textbf{44.5} &43.0 &\textbf{60.9} &26.5 &43.2 &\textbf{43.4} &27.2 &37.4 &\textbf{40.8}\\
\hline
Supervised target & &47.4 &40.8 &66.8 &27.2 &48.2 &32.4 &31.2 &38.3 &41.5\\
\hline
EPM~\cite{EPM}& &41.5 &43.6 &57.1 &\textbf{29.4} &\textbf{44.9} &\textbf{39.7} &29.0 &36.1 &40.2\\
FCOS (source-only) & &33.8 &34.8 &39.6 &18.6 &27.9 &6.3 &18.2 &25.5 &25.6\\
\emph{DC-only} FCOS &ResNet-101 & 39.4 &41.1 &54.6 &23.8 &42.5 &31.2 &25.1 &35.1 &36.6\\
Ours & &\textbf{43.6} &\textbf{46.7} &\textbf{62.1} &27.8 &44.0 &37.0 &\textbf{29.9} &\textbf{38.4} &\textbf{41.2}\\
\hline
Supervised target & &44.7 &43.9 &64.7 &31.5 &48.8 &44.0 &31.0 &36.7 & 43.2\\
\hline
\end{tabular}}

\label{table:val}
\end{table*}

\section{Experiments}
\label{Experiments}
We investigate the effectiveness of our \OurMethod{} by carrying out experiments on  scenarios of various domain shifts: 1. weather variation, 2. cross-camera adaptation, 3. synthetic-to-real discrepancy. Our~\OurMethod{} outperforms the UDA Object Detection state-of-the-arts on different backbones. We also provide abundant ablation experiments to prove the robustness of our method. 
For different backbone, e.g. VGG-16~\cite{vgg16} and ResNet-101~\cite{res101}, the optimal structure of multi-level alignments is explored  
\subsection{Datasets}
Four public datasets are applied in our experiments, including Cityscapes~\cite{cityscape}, Foggy Cityscapes~\cite{foggy-cityscape}, Sim10k~\cite{sim10k}, and KITTI~\cite{kitti}. 
In all experiments, we train on source images with ground truth labels and on target images without annotation. We introduce the evaluation scenarios of various domain shifts in the following.

\textbf{Weather variation} We choose Cityscapes~\cite{cityscape} as the source domain and Foggy Cityscapes~\cite{foggy-cityscape} as the target domain. Cityscapes is a city street scene dataset collected from different cities in Germany. It contains 5000 images (2975 in training set and 500 in validation set) with eight-class detection labels. Foggy Cityscapes is a synthetic dataset rendered from the Cityscapes dataset. By adding fog noise of three density levels on each image in Cityscapes separately, Foggy cityscapes simulates scenarios of driving in a foggy wheather. It contains 8925 training images and 1500 validation images. We evaluate our approach on the validation set of Foggy Cityscapes with all eight categories.

\textbf{Cross-camera adaptation}  We use KITTI~\cite{kitti} and Cityscapes~\cite{cityscape} as the source and the target domain. KITTI is one of the most popular datasets in autonomous driving, which contains 7481 images with annotations in the training set. Images in KITTI are collected by a driving car in urban, rural and highway scenes. Compared with Cityscapes, KITTI has a different camera setup. Following the setting in other works~\cite{da_frcnn,EPM}, we only evaluate our approach on the Cityscapes validation set with the car category. 

\textbf{Synthetic-to-real-world adaptation}  
SIM10K is a collection of synthesized driving scene images generated by the grand theft auto (GTAV) engine. It contains 10000 images with bounding box annotations. We evaluate our~\OurMethod{} adaptation ability from synthesized to real-world images by applying all images of SIM10K as source domain training set while Cityscapes as the target domain.
Considering that only objects of the car category are annotated in SIM10K, we perform evaluation only on the car category.
\subsection{Implementation details}
Following the hyper-parameter setting in ~\cite{EPM}, in all experiments, we set the $\lambda$ and $\beta$ in Eq.~\ref{eq:5} to 0.1 and 1.0 and set the reversed gradient weight in GRL to 0.01. We further choose FCOS~\cite{FCOS} as our detector. By default, we utilize VGG-16~\cite{vgg16} or ResNet-101~\cite{res101} as our backbone and initialize parameters of them by the pre-trained module on ImageNet~\cite{ImageNet}. All networks are trained with the base learning rate of $5 \times 10^{-3}$ for 20K iterations, the constant warm-up of 500 iterations, the momentum of 0.9, and the weight decay of $5 \times 10^{-4}$.

\subsection{Main results}
We compare our \OurMethod{} with previous state-of-the-art approaches including DA Faster R-CNN series and one-stage methods. We present three baselines: two-stage Faster R-CNN, FCOS (our object detector) and \emph{DC-only} FCOS. The first two are trained only on the source domain without adaptation. \emph{DC-only} FCOS has the same domain discriminator architecture as our multi-level alignment module without AIR branches. We also present the supervised target results where FCOS is trained with ground truth labeled target data. By default, we use $\mathbf H_3, \mathbf H_4$ for multi-level AIR on the VGG-16 model, and $\{ \mathbf H_3,\mathbf H_4,\mathbf H^f_5,\mathbf H^f_6,\mathbf H^f_7\}$ on the ResNet-101 model. Moreover, we apply domain discriminators on feature maps from $ \mathbf P_3$ to $\mathbf P_7$ in all experiments.

\begin{table}[t!]
\centering
\small
\caption{ \footnotesize Quantitative results for KITTI $\rightarrow$ Cityscapes dataset. Results of each category are evaluated on mAP$^r_{0.5}$
}
\begin{tabular}{rc|c}
\hline
Method & Backbone & mAP$^r_{0.5}$\\
\hline
\hline
Faster R-CNN (source-only)& &30.2 \\
DA-Faster~\cite{da_frcnn}& &38.5\\ 
SW-DA~\cite{SW-DA}& &42.3 \\
MAF~\cite{MAF}& &41.0\\
ATF~\cite{ATF}&VGG-16 & 37.9\\
SC-DA~\cite{SC-DA}& &42.5\\
EPM~\cite{EPM}& &43.2\\
FCOS(source-only)& &34.4\\
\emph{DC-only} FCOS& &39.1\\
Ours& &\textbf{45.2}\\
\hline
Supervised target& & 69.7\\
\hline
EPM~\cite{EPM}& &45.0 \\
FCOS (source-only) & &35.3\\
\emph{DC-only} FCOS &ResNet-101 &42.3\\
Ours& &\textbf{45.3}\\
\hline
Supervised target& &  70.4 \\
\hline
\end{tabular}

\label{table:kitti}
\end{table}

\begin{figure*}[htb]
\includegraphics[scale=0.73]{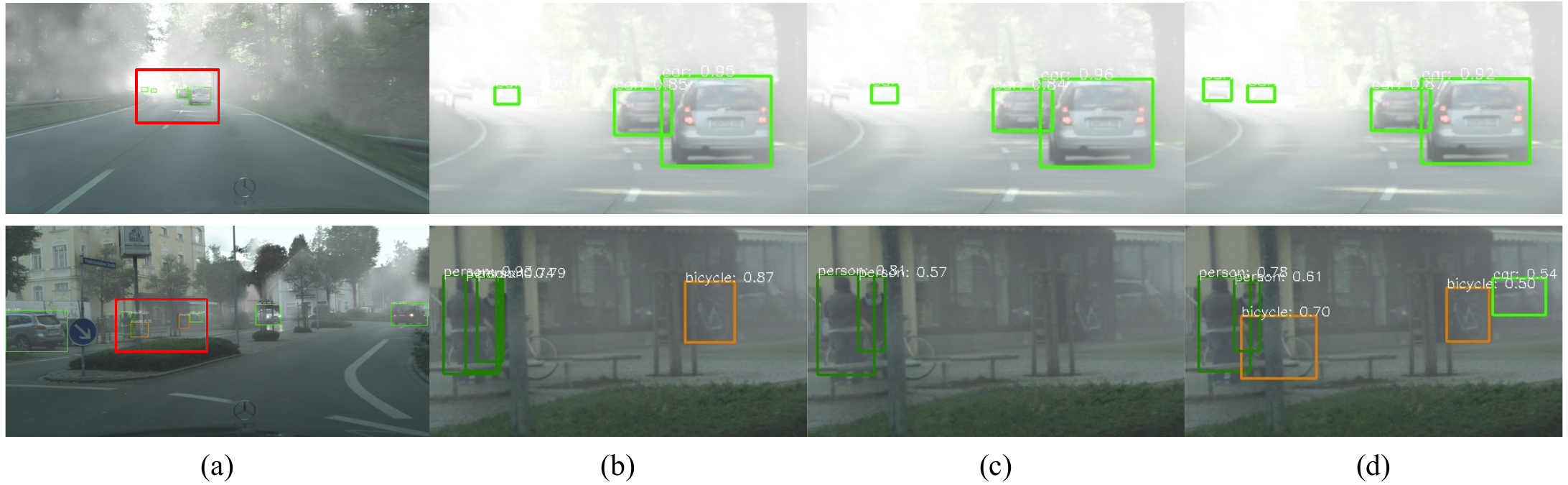}
\centering
\caption{\textbf{Foggy-detail.} We compare the detection results of different approaches on two image samples (top and bottom row). We illustrate results on the entire image by AIR-DA (subfigure (a)), and results on a zoomed-in patch (of the red box) by DC-only FCOS (subfigure (b)), EPM (subfigure (c)) and AIR-DA (subfigure (d)) for clear comparison. }
\label{fig:foggy_detail}
\end{figure*}

\textbf{Weather variation}
As shown in Table~\ref{table:val}, source-only FCOS has worse performance than source-only Faster R-CNN. However, our \OurMethod{} outperforms existing FRCNN-based approaches and beats Prior-DA~\cite{Prior-DA} by 1.5$\%$ mAP. When compared to EPM~\cite{EPM}, the best one-stage method among most recent approaches, \OurMethod{} surpasses by 4.8$\%$ mAP on the VGG-16 backbone, and by 1.0$\%$ mAP on the ResNet-101 backbone. Moreover, our method is more effective than EPM which has two training sessions. Compared with the \emph{DC-only} FCOS, the proposed AIR branch can significantly improve the performance by 7.6$\%$ mAP with the VGG-16 backbone. We also give qualitative comparison in Figure~\ref{fig:foggy_detail}. We illustrate the detection results of AIR-DA in the entire image on the left side. We further capture images in the red box and zoom in our result to compare with \emph{DC-only} FCOS and EPM~\cite{EPM}. From the top row, we notice that compared with other methods, our \OurMethod~ predicts with fewer false positives especially for distant objects. From the second row, we find that~\OurMethod~also works better on occluded objects, e.g. the bicycle behind the road signs and the occluded people.
\begin{figure*}[htb]
\includegraphics[scale=0.73]{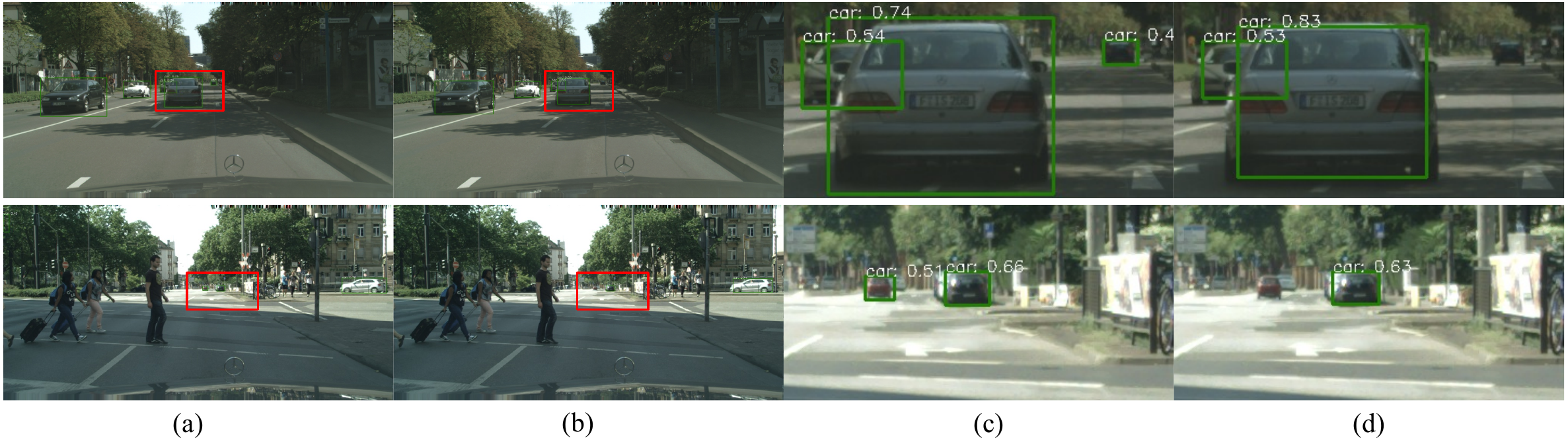}
\centering
\caption{\textbf{KITTI-detail.} 
We compare the detection results of AIR-DA and EPM on two image samples (top and bottom row). We illustrate results on the entire image by AIR-DA (subfigure (a)) and EPM (subfigure (b)) separately. We further zoom in both results in the red box (AIR-DA (subfigure (c)) and EPM (subfigure (d))) for clear comparison.}
\label{fig:Kitti_detail}
\end{figure*}

\textbf{Cross-camera adaptation}
We report the results of cross-camera adaptation in Table~\ref{table:kitti}. For both backbones of VGG-16 and ResNet-101, our \OurMethod{} can achieve the best performance. Compared with FCOS without adaptation, the proposed method, taking advantage of AIR branch and multi-level alignments module, brings a drastic boost in performance by more than 10$\%$ mAP. Moreover, even though introducing a new adversarial task, our framework is also label-free on the target domain. Qualitative comparison can be found in Figure~\ref{fig:Kitti_detail}. Specifically, the left two columns demonstrate the detection results on the entire image by AIR-DA and EPM. The right two columns are the zoomed-in results for more clear visualization. From both rows, we observe that the proposed method is competent on detection of distant objects in the image which are blurred and difficult for \cite{EPM}. 

\begin{table}[h]
\centering
\small
\caption{ \footnotesize Quantitative results for SIM10K $\rightarrow$ Cityscapes dataset. Results of each category are evaluated on mAP$^r_{0.5}$
}
\begin{tabular}{rc|c}
\hline
Method & Backbone & mAP$^r_{0.5}$\\
\hline
\hline
Faster R-CNN (source-only)& &30.1 \\
DA-Faster~\cite{da_frcnn}& &38.9\\ 
SW-DA~\cite{SW-DA}& &42.3 \\
MAF~\cite{MAF}& &41.1\\
ATF~\cite{ATF}&VGG-16 &42.8\\
SC-DA~\cite{SC-DA}& &43.0\\
RPA~\cite{RPA}& &45.7\\
EPM~\cite{EPM}& &49.0\\
FCOS(source-only)& &39.8\\
\emph{DC-only} FCOS& &45.9\\
Ours& &\textbf{49.4}\\
\hline
Supervised target& & 69.7\\
\hline
EPM~\cite{EPM}& &51.2 \\
FCOS(source-only)& &41.8\\
\emph{DC-only} FCOS & ResNet-101 &50.6\\
Ours& &\textbf{53.3}\\
\hline
Supervised target& &  70.4 \\
\hline

\end{tabular}

\label{table:sim10k}
\end{table}

\begin{figure*}[htb]
\includegraphics[scale=0.73]{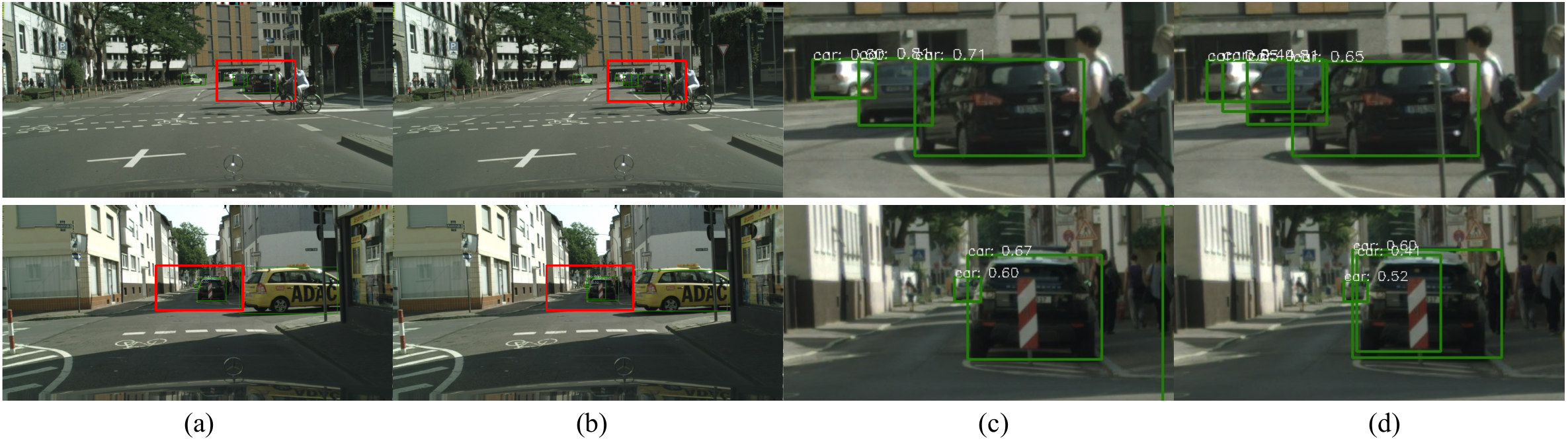}
\centering
\caption{\textbf{SIM10K-detail.} 
We compare the detection results of AIR-DA and EPM on two image samples (top and bottom row). We illustrate results on the entire image by AIR-DA (subfigure (a)) and EPM (subfigure (b)) separately. We further zoom in both results in the red box (AIR-DA (subfigure (c)) and EPM (subfigure (d))) for clear comparison.}
\label{fig:sim10k_detail}
\end{figure*}
\textbf{Synthetic-to-real-world adaptation}
In Table~\ref{table:sim10k}, we report the results of domain adaptation from SIM10K synthetic data to Cityscapes real-world street scene. We observe that our method sets new records for UDA Object Detection, outperforming the recent best method on both backbones of ResNet-101 and VGG-16. Our multi-level AIR module also surpasses \emph{DC-only} FCOS by 3.5$\%$ mAP with VGG-16 backbone and 2.7$\%$ mAP with ResNet-101 backbone. We further give qualitative comparison in Figure~\ref{fig:sim10k_detail}. Consistent with the conclusions drawn from other domain adaptation scenarios, those quantitative comparisons and qualitative detection results demonstrate that the proposed AIR module can achieve better feature alignment performance, especially for low-level features.

\subsection{Ablation experiments and discussion}
We study the effectiveness of our multi-level adversarial feature alignment module by carrying out ablation experiments and intuitively visualizing the feature adaptation. Notably, in all experiments, we use multi-level domain discriminator with FPN from $\mathbf{F_3}$ to $\mathbf{F_7}$. We also discuss the model selection of AIR with experiments on different backbones.
\begin{table}[h]
\centering
\small
\caption{ \footnotesize \textbf{Reconstruction feature locations}: Performance with VGG-16 backbone for Cityscapes $\rightarrow$ Foggy-Cityscapes dataset. $\mathbf P_3,\mathbf P_4$ uses features from FPN. $\mathbf H_3, \mathbf H_4$ uses features from domain tower.}
\label{table:H34}
\begin{tabular}{r|ccc}
\hline
Features&mAP &mAP$^r_{0.5}$ &mAP$^r_{0.75}$\\
\hline
\hline
$\mathbf P_3, \mathbf P_4$ &20.5 & 40.3 &18.1\\
$\mathbf H_3, \mathbf H_4$ &\textbf{21.4} &\textbf{40.8} &\textbf{19.9}\\
\hline
\end{tabular}

\end{table}

\textbf{AIR feature locations: FPN} \textit{vs.} \textbf{domain tower} We compare our AIR feature sampling locations. As shown in Table~\ref{table:H34}, by using the domain tower features $\mathbf{H}$, we can improve the performance in various mAP metrics w.r.t different IoU thresholds, e.g., 0.9$\%$~in mAP, 0.5$\%$ in mAP$^r_{0.5}$ and 1.8$\%$ in mAP$^r_{0.75}$ respectively. In the following experiments, if not stated otherwise, we use output of the domain tower as our AIR branch input.

\begin{table}[h]
\centering
\small
\caption{ \footnotesize \textbf{AIR module with VGG-16 backbone}: Performance of multi-level AIR module for Cityscapes $\rightarrow$ Foggy-Cityscapes dataset. `+ $\mathbf H_3$' only take single-level AIR, using $\mathbf H_3$ features. `+ $\mathbf H_4$' applies two-level module by  $\mathbf H_3$ and $\mathbf H_4$ features. Notably, `+ $\mathbf H_5$' uses $\mathbf H_3,\mathbf H_4,\mathbf H_5$ features while `+ $\mathbf H^f_5$' uses $\mathbf H_3,\mathbf H_4,\mathbf H^f_5$ instead.}
\label{table:AM_VGG16}
\begin{tabular}{r|c|ccc}
\hline
Alignments &Backbone &mAP &mAP$^r_{0.5}$ &mAP$^r_{0.75}$\\
\hline
\hline
\emph{DC-only} & &18.0 &33.2 &16.5\\
+ $\mathbf H_3$ &  &20.5 &38.8 &18.5\\
+ $\mathbf H_4$&VGG-16 &\textbf{21.4} &\textbf{40.8} &\textbf{19.9}\\
+ $\mathbf H_5$& & 20.1& 38.9& 18.8\\
+ $\mathbf H^f_5$& & 20.2& 39.2& 18.8\\
\hline
\end{tabular}

\end{table}

\textbf{Multi-level AIR module with shallow backbone}: In Table~\ref{table:AM_VGG16}, we report our results with different AIR modules on the VGG-16 backbone and show that, with our AIR, the adaptation performance can be further improved, even if just applied on features of a single level. Since VGG-16 is a shallow backbone, we find that using two-level AIR with $\mathbf H_3$ and $\mathbf H_4$ leads to the best performance. Therefore, we apply two-level AIR with $\mathbf H_3$ and $\mathbf H_4$ in all experiments with the backbone of VGG-16. We also notice that using $\mathbf H^f_5$ instead of the original $\mathbf H_5$ can improve the performance.

\begin{figure}[h]
    \centering
    \includegraphics[scale=0.15]{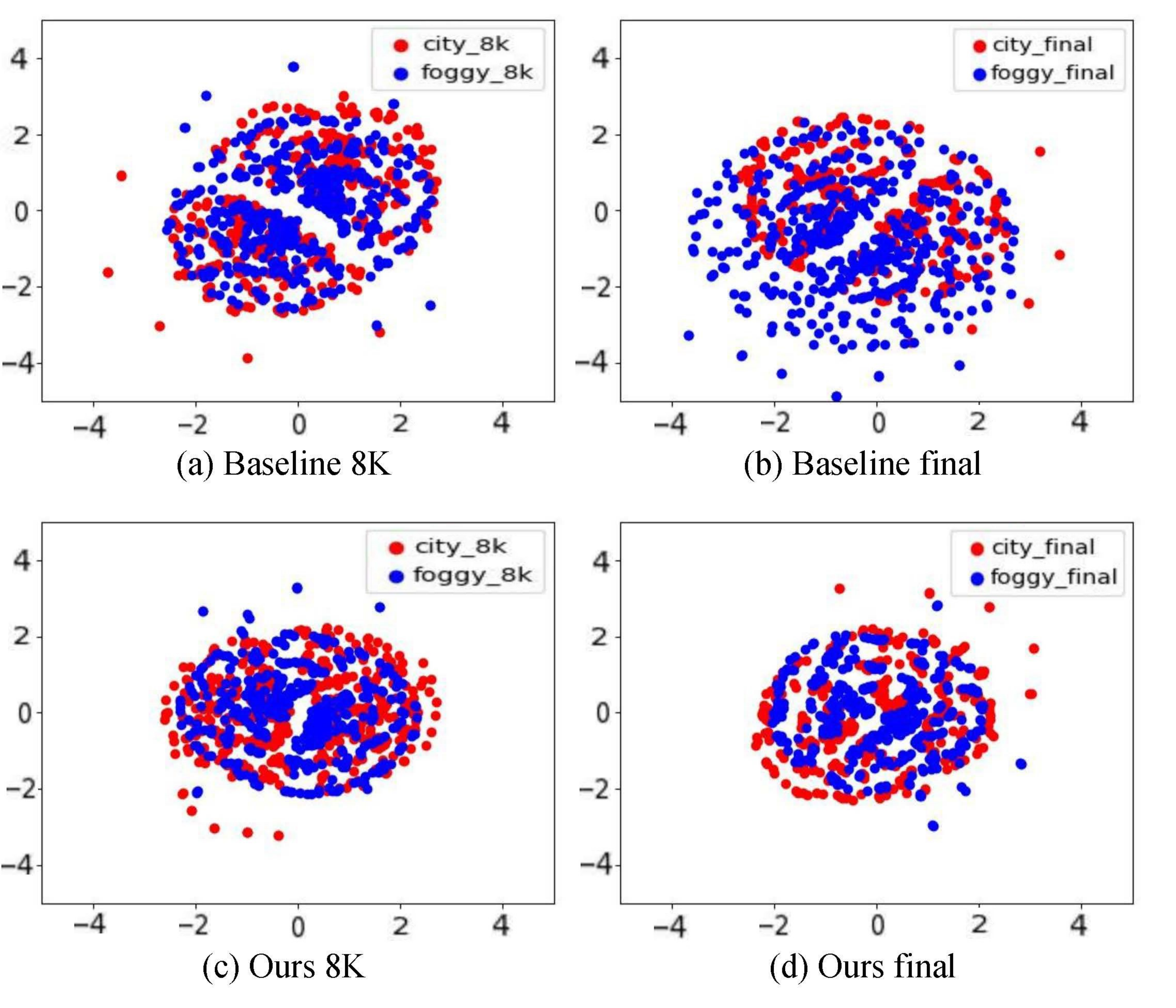}
     \caption{ \footnotesize Visualization of FPN features with t-SNE~\cite{t-SNE}. The red points represent source examples from Cityscape while the blue ones represent target samples from Foggy Cityscape. All figures visualize features of `P3’ layer in FPN. Subfigure (a) and (b) are features at the 8K-th step and at the last step, with adaptation adaptation by our baseline which only uses a domain discriminator for multi-level adversarial feature alignments. Subfigure (c) and (d) are corresponding features with adaptation by our multi-level adversarial feature alignments which contain a domain discriminator as well as the AIR module. Compared to the baseline, our method can better align features and maintain the good alignment until the convergence of the detector training is reached.}
\label{figure:visual}
\end{figure}

\begin{table}[h]
\centering
\small
\caption{ \footnotesize \textbf{AIR module with ResNet-101 backbone}: Performance of multi-level AIR module for Cityscapes $\rightarrow$ Foggy-Cityscapes dataset. $\mathbf H_3, \mathbf H_4$ uses $\mathbf H_3$ and $\mathbf H_4$ features for AIR. `+$\mathbf H^f_5$' uses $\mathbf H_3,\mathbf H_4,\mathbf H^f_5$ features and so on. `+$\mathbf H_5 \sim \mathbf H_7$' applies all original $H_j$ features. `+ different $\mathbf H^f_j$' means $\mathbf H_5,\mathbf H_6, \mathbf H_7$ fuses $\mathbf C_2,\mathbf C_3,\mathbf C_4$ respectively.}
\begin{tabular}{r|c|ccc}
\hline
Alignments &Backbone &mAP &mAP$^r_{0.5}$ &mAP$^r_{0.75}$\\
\hline
\hline
$\mathbf H_3, \mathbf H_4$ &  &20.7 &40.7 &19.5\\
+ $\mathbf H^f_5$ & &21.0 &41.0 &20.0\\
+ $\mathbf H^f_5\sim H^f_7$&ResNet-101 & 21.4& \textbf{41.2} & \textbf{20.5}\\
+ $\mathbf H_5 \sim \mathbf H_7$& & 21.0& 41.0& 20.3\\
+ different $\mathbf H^f_j$  & & \textbf{22.0} & 41.0& \textbf{20.5}\\
\hline
\end{tabular}

\label{table:AM_R101}
\end{table}

\textbf{Multi-level AIR module with deeper backbone}: For a deeper backbone like ResNet-101, we evaluate our domain adaptive detector with different multi-level AIR modules. As shown in Table~\ref{table:AM_R101}, the two-level AIR module, which yields best results on the VGG-16 backbone, leads to suboptimal performance. Notice that even using all $\mathbf H_j$ features can get comparable results, fusing backbone features can improve the results in all mAP metrics. Moreover, other fusion methods( $\mathbf H_5,\mathbf H_6, \mathbf H_7$ fuses $\mathbf C_2,\mathbf C_3,\mathbf C_4$ respectively) should also work without much difference. It should be noted that fusion does not introduce computation overhead in testing, we choose `+$\mathbf H^f_5\sim \mathbf H^f_7$' in all our experiments with ResNet-101 backbone.

\textbf{Feature visualization and qualitative analysis}: We further visualize the FPN features learned for domain adaptation on Cityscapes $\rightarrow$ Foggy-Cityscapes, using t-SNE~\cite{t-SNE}. For ease of visualization, we only display $\mathbf P_3$ features obtained by two methods, \emph{DC-only} baseline and our AIR module. From the log of training loss, we observe that adversarial domain discrimination loss almost converges to zero at about 8K iterations but detection loss declines constantly. Therefore, we plot the features both at the 8K-th and at the last iteration.  

Illustrated in Figure~\ref{figure:visual}, the blue points represent target samples while the red ones represent source examples. At the 8K-th iteration, the domain discriminator has been well trained and both AIR-DA and \emph{DC-only} methods achieve good feature alignments. However, although detector work better when continuing to train DA detector, the well-trained domain discriminator fails to maintain adversarial training balance, making shift in final results. Compared with this \emph{DC-only} method, our AIR module can keep better adversarial training balance and get better feature alignment performance in the final. 

\section{Conclusion}
In this paper, we devise a simple and novel  Adversarial Image Reconstruction (AIR) branch for the domain adaptive object detection task. Our AIR module helps the domain adaptive network to achieve better balance of adversarial training between the feature extractor and the domain discriminator, by encouraging the feature extractor to disregard fine-grained details. We further design multi-level feature alignments for adaptation on different FPN layers. 
Combining the AIR branch and the multi-level alignment pipeline, we construct the \OurMethod{} framework on the one-stage detection FCOS. We demonstrate that AIR-DA maintains satisfying feature alignments throughout the training session of the detector.
 Our \OurMethod~demonstrates efficacy on multiple domain adaptation scenarios of various domain shifts, and outperforms almost all state-of-the-art predecessors of both one- and two-stage architectures.
\bibliographystyle{IEEEtran}
\bibliography{ref}
\end{document}